\documentclass[lettersize,journal]{IEEEtran}
\usepackage{amsmath,amsfonts}
\usepackage{algorithmic}
\usepackage{algorithm}
\usepackage{array}
\usepackage[caption=false,font=normalsize,labelfont=sf,textfont=sf]{subfig}
\usepackage{textcomp}
\usepackage{stfloats}
\usepackage{url}
\usepackage{verbatim}
\usepackage{graphicx}
\usepackage{cite}

\usepackage{lineno}
\usepackage[colorlinks]{hyperref}
\usepackage{xcolor}
\usepackage{mathtools}
\usepackage{amsthm}
\usepackage{amssymb}
\usepackage{dsfont}
\usepackage{subcaption}

\usepackage{mathrsfs}
\usepackage{arydshln}
\usepackage{multirow}
\usepackage{booktabs}
\usepackage{microtype}

\theoremstyle{plain}
\newtheorem{theorem}{Theorem}[section]
\newtheorem{proposition}[theorem]{Proposition}

\theoremstyle{definition}

\theoremstyle{remark}
\newtheorem{remark}[theorem]{Remark}
\begin{document}

\title{Information-Theoretic Decoupled Prompt Tuning for Continual Learning}

\author{Yunfei Zhang, Wen Wen, Tieliang Gong, and Weizhan Zhang,~\IEEEmembership{Senior Member,~IEEE}
\thanks{This work was supported in part by the National Natural Science Foundation of China under Grant 62576268 and in part by the Fundamental Research Funds for the Central Universities under Grant xxj032025002.}
\thanks{(Yunfei Zhang and Wen Wen contributed equally to this work.)(Corresponding author: Tieliang Gong.)}
\thanks{The authors are with the National Engineering Lab for Big Data Analytics, School of Computer Science and Technology, Xi’an Jiaotong University, Xi’an 710049, China (e-mail: cloudfly.zyf@gmail.com; wen190329@gmail.com; adidasgtl@gmail.com; zhangwzh@xjtu.edu.cn).}
}

\markboth{IEEE Transactions on Multimedia}%
{Zhang \MakeLowercase{\textit{et al.}}: Information-Theoretic Decoupled Prompt Tuning for Continual Learning}


\maketitle

\begin{abstract}
Continual learning (CL) aims to incrementally acquire knowledge from sequential data while avoiding catastrophic forgetting. Recently, prompt tuning has attracted increasing attention as an efficient approach for adapting pre-trained models to CL tasks. However, existing prompt design paradigms commonly suffer from retrieval dependence and classifier bias, which make model adaptation sensitive to prompt selection and bias predictions toward newly arrived classes. To address these challenges, we propose \textbf{D}ecoupled \textbf{P}rompt \textbf{T}uning for \textbf{C}ontinual \textbf{L}earning (DPT4CL), which decouples the CLIP textual prompt into a task-shared prompt distribution and class-specific prompts. The task-shared prompt distribution is derived by optimizing an Information Bottleneck objective to facilitate cross-task knowledge transfer and alleviate classifier bias, while class-specific prompts enhance inter-class separability without relying on explicit prompt retrieval. Furthermore, we establish a unified excess risk bound from an information-theoretic perspective, providing theoretical support for the robust generalization and forgetting mitigation of the proposed framework. Extensive experiments on standard CL benchmarks demonstrate that DPT4CL achieves state-of-the-art performance. The source code is available at \url{https://github.com/Cloudfly-Z/DPT4CL}.
\end{abstract}


\section{Introduction}
\IEEEPARstart{C}{ontinual} learning (CL) aims to incrementally acquire knowledge from non-stationary task streams without degrading performance on previously learned tasks \cite{shi2025continual}. Among diverse CL scenarios, class-incremental learning poses the most significant challenge, as it requires the learner to distinguish among all encountered classes without access to task identities as new classes arrive sequentially \cite{zhou2024class}. Given the constraints on resources and privacy in the real world, retaining or accessing all previously observed data is often infeasible \cite{chaudhry2019tiny}. This inevitably leads to \textit{catastrophic forgetting} \cite{kirkpatrick2017overcoming}, where the learner adapts to new tasks at the expense of overwriting the essential knowledge preserved for previous ones. 

Recently, large-scale pre-trained models such as ViT \cite{dosovitskiy2021an} and CLIP \cite{radford2021learning} have demonstrated strong generalization ability across diverse applications \cite{li2023lift,jha2024clap4clip}. To adapt these foundation models efficiently, prompt tuning has emerged as a lightweight parameter-efficient fine-tuning paradigm, introducing far fewer trainable parameters than other parameter-efficient methods, such as Adapters \cite{chen2022adapterformer} and LoRA \cite{hu2022lora}. This property makes it well suited to continual learning, where models need to incrementally acquire new task knowledge with minimal parameter growth. For instance, recent prompt pool methods, including L2P \cite{wang2022learning}, DualPrompt \cite{wang2022dualprompt}, CODA-Prompt \cite{smith2023coda}, and AttriCLIP \cite{wang2023attriclip}, freeze the pre-trained parameters and optimize a prompt pool composed of key--prompt pairs. During inference, they retrieve the top-$N$ prompts for each test instance via key--query matching. However, their performance is sensitive to the pool capacity and the choice of $N$, which may lead to retrieving sub-optimal prompts. Moreover, their focus on task-specific knowledge may overlook correlations among sequential task distributions.

\begin{figure}[t]
\centering
\includegraphics[width=\columnwidth]{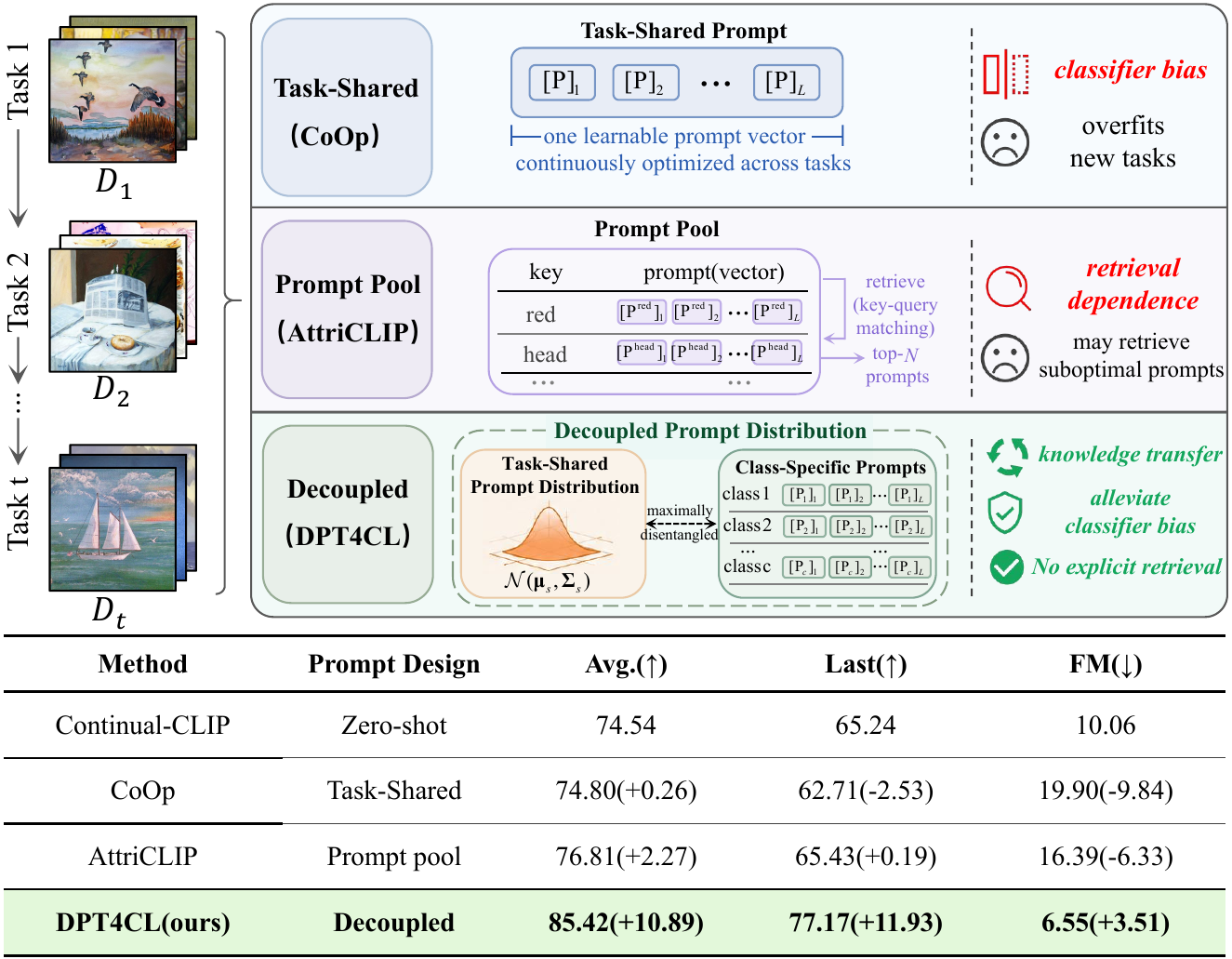}
\caption{
Different prompt design paradigms for continual learning and their average performance comparison across four datasets.
Continual-CLIP serves as the zero-shot recognition baseline of CLIP in continual learning, and the signed values denote performance improvements relative to this baseline.
}
\label{fig:prompt_design}
\vspace{-5mm}
\end{figure}

\IEEEpubidadjcol
An alternative is to learn task-shared knowledge. For example, CoOp \cite{zhou2022learning} continuously optimizes shared learnable prompts in the textual branch of CLIP, avoiding explicit prompt retrieval during inference. Nevertheless, without past task data, this strategy is prone to severe \textit{classifier bias} \cite{chen2025adaptive,wu2019large}: the shared prompts tend to overfit newly arrived tasks and misclassify past task samples into novel classes. Overall, existing prompt-based methods still suffer from retrieval dependence or classifier bias, limiting their potential to mitigate catastrophic forgetting in real-world CL settings. To address these challenges, we propose Decoupled Prompt Tuning for Continual Learning (DPT4CL), which decouples the prompt distribution into task-shared and class-specific components from an information-theoretic perspective, as shown in Fig. \ref{fig:prompt_design}. This prompt design facilitates knowledge transfer across tasks while avoiding explicit prompt retrieval and alleviating classifier bias. The proposed DPT4CL consistently outperforms the zero-shot recognition baseline, Continual-CLIP \cite{thengane2022clip}, and existing prompt-based CL methods such as CoOp and AttriCLIP across the four datasets.

Specifically, DPT4CL decomposes the CLIP textual prompt space into two complementary components: a \textit{task-shared prompt distribution}  and \textit{class-specific prompts}. The task-shared prompt distribution captures shared knowledge across sequential tasks. To this end, we leverage the Information Bottleneck (IB) principle to learn a compact yet sufficient unified prompt distribution by maximizing the mutual information (MI) between the prompt and class labels while minimizing the MI between input images and the prompt. This suppresses task-irrelevant visual redundancy while preserving discriminative task-shared semantics. We further introduce a distillation objective consisting of prompt-level distributional distillation and feature-level language-aware distillation to alleviate the classifier bias observed in CoOp-like prompt optimization. In parallel, class-specific prompts are introduced to encode category-specific semantics and enhance inter-class separability. Owing to the distinctive textual architecture of CLIP, each class-specific prompt can be directly associated with the [CLS] token of its corresponding category, avoiding task-specific prompt retrieval during inference. An orthogonalization objective is then employed to maximally disentangle class-specific prompts from the task-shared prompt. In this way, the task-shared prompt promotes stable cross-task knowledge transfer, while class-specific prompts strengthen local category discrimination, enabling DPT4CL to achieve state-of-the-art performance. Our main contributions are summarized as follows:
\begin{itemize}
    \item We propose DPT4CL, a textual prompt tuning framework that decouples the CLIP textual prompt into a task-shared prompt distribution and class-specific prompts. The task-shared prompt distribution facilitates cross-task knowledge transfer, while distillation preserves semantic consistency to alleviate classifier bias. Meanwhile, class-specific prompts enhance inter-class separability through orthogonal decoupling from the task-shared prompt.
        
    \item We establish a unified excess risk bound through the lens of information theory, decomposing the overall risk into the \textit{Empirical Suboptimality Gap}, \textit{Generalization Gap}, and \textit{Forgetting}. Our theoretical results formally demonstrate that the proposed decoupled prompt framework serves as a computationally tractable surrogate for controlling these critical quantities, thereby synergistically mitigating catastrophic forgetting and improving robust generalization.

    \item Extensive experiments on standard CL benchmarks demonstrate the effectiveness of DPT4CL. The proposed framework consistently outperforms strong zero-shot CLIP baselines and existing state-of-the-art CL methods.
\end{itemize}

\section{Related Work}
\subsection{Continual learning}
Existing continual learning scenarios can be broadly categorized into three settings \cite{shi2025continual}: task-incremental learning, where task identities are available during both training and inference \cite{masana2021ternary,hossain2022rethinking}; domain-incremental learning, where data distribution shifts while the task structure remains constant \cite{shi2023unified,luo2026domain}; and class-incremental learning, where new classes sequentially arrive and require the model to distinguish between all encountered classes without access to task identifiers \cite{belouadah2021comprehensive}. Among these, class-incremental learning is the most challenging as it necessitates establishing a global decision boundary across all encountered classes, leading to inter-task interference and catastrophic forgetting. This has led to a proliferation of work aimed at designing classifiers to alleviate forgetting through functional regularization that penalizes critical changes to previous tasks \cite{sun2023regularizing,li2024continual}, modularized architectures that isolate task parameters \cite{douillard2022dytox,liu2025lora}, and rehearsal strategies that retain data from previous tasks for joint training with the current task \cite{rolnick2019experience,smith2024adaptive}. Despite substantial progress, the performance of these methods often relies heavily on parameter-intensive modular expansion or the capacity of rehearsal buffers. In this paper, we address class-incremental learning with lightweight soft prompt tuning in a rehearsal-free setting, enabling effective use of pre-trained models while avoiding excessive parameter growth and memory dependence.

\subsection{Prompt-based continual learning}
Recent extensive work has harnessed the powerful generalizability of pre-trained models by integrating them as backbones into CL frameworks \cite{thengane2022clip,liu2025c}, thereby enhancing the model's discriminative capability. However, in practice, their zero-shot performance often degrades on downstream out-of-domain data \cite{wortsman2022robust,pham2023combined}. Prompt tuning has emerged as a prevalent paradigm for efficient adaptation, which leverages learnable continuous tokens as inputs to the frozen visual and/or textual encoders to capture task-specific knowledge \cite{zhou2022learning,zhou2022conditional}. For example, inspired by VPT \cite{jia2022visual}, a series of vision-only prompt tuning CL approaches have been proposed. Among them, L2P \cite{wang2022learning} maintains a learnable visual prompt pool and utilizes a key-query matching strategy to dynamically retrieve instance-specific prompts, thereby effectively adapting to sequential tasks. DualPrompt \cite{wang2022dualprompt} and CODA-Prompt \cite{smith2023coda} further advance this approach by incorporating complementary visual prompts into the pre-trained ViT backbone to capture rich task-specific information, thereby boosting performance. Meanwhile, benefiting from the rich semantic guidance provided by CLIP, an alternative line of textual soft prompt tuning approaches has emerged. For instance, methods such as CoOp \cite{zhou2022learning} and AttriCLIP \cite{wang2023attriclip} adapt to CL scenarios by encapsulating task-relevant knowledge within textual prompts. Specifically, they employ learnable prompt vectors or prompt pool to extract linguistic semantics complementary to visual representations, thereby achieving favorable generalization across sequential tasks. Nevertheless, both vision-only prompt tuning and textual soft prompt tuning methods still suffer from the challenges of retrieval dependence or classifier bias, as discussed in the Introduction. In contrast to these approaches, we decouple the prompt distribution into task-shared and class-specific components from an information-theoretic perspective. This design facilitates effective transfer of task-shared knowledge while eliminating the dependence on explicit prompt retrieval during inference.

\section{Methodology}
\subsection{Preliminaries}
\noindent\textbf{Continual learning (CL).} Class-incremental CL aims to continuously learn knowledge of new classes from $T$ sequential tasks $\{D_1,\ldots, D_T\}$ without forgetting knowledge of old classes. For each task $t \in [T]$, the dataset $D_t = \{(\mathbf{x}_i^t, y_i^t)\}_{i=1}^{n_t}$ comprises $n_t$ pairs of input samples $\mathbf{x}_i^t \in \mathcal{X}$ and their corresponding labels $y_i^t \in \mathcal{C}_t$, drawn i.i.d. from the underlying data distribution $\mathcal{D}_t$ over the joint space $\mathcal{X} \times \mathcal{C}_t$, where $\mathcal{X}$ denotes the feature space and $\mathcal{C}_t$ denotes the class space for task $t$. For different tasks, the class spaces are typically disjoint, i.e., $\forall i,j\in[T]$ and $i\neq j$, $\mathcal{C}_i \cap \mathcal{C}_j = \emptyset$. When the $t$-th task arrives, the learner accesses $D_t$ together with limited samples (if any) from previous tasks $\{D_1, \ldots, D_{t-1}\}$, with the objective of classifying samples over all observed classes $\mathcal{C} = \bigcup_{i=1}^t \mathcal{C}_i$.

\noindent\textbf{CLIP-based prompt tuning.} CLIP \cite{radford2021learning} contains an image encoder $f_{v}(\cdot)$ and a text encoder $f_t(\cdot)$, which are jointly pre-trained on large-scale image-text pairs via a contrastive learning objective. Specifically, an image $\mathbf{x}$ and a text $\mathbf{t}$ are processed by encoders $f_{v}(\cdot)$ and $f_t(\cdot)$ to obtain the corresponding feature representations $\mathbf{z}_v$ and $\mathbf{z}_t$. Here, $\mathbf{t}$ is the sequence of text tokens derived from a prompt template such as the hand-crafted prompt  ``A photo of a [CLS]'', where $\text{[CLS]}$ denotes a specific class label. By replacing $\text{[CLS]}$ with each class name, we can obtain the set of text inputs $\{\mathbf{t}_c\}_{c=1}^C$, where $C$ denotes the total number of all observed classes. These inputs are then processed by the encoder $f_t(\cdot)$ to generate the corresponding representations $\{\mathbf{z}_{t,c}\}_{c=1}^C$. Consequently, the prediction probability for image $\mathbf{x}$ belonging to the $i$-th class $y_i$ can be computed by:
\begin{equation}
    p(y_i|\mathbf{x}) = \frac{\exp(\langle \mathbf{z}_v, \mathbf{z}_{t,y_i}\rangle/\tau)}{\sum_{c=1}^{\vert C \vert} \exp (\langle\mathbf{z}_v, \mathbf{z}_{t,c} \rangle/\tau)},
\end{equation}
where $\langle \cdot,\cdot \rangle$ is the cosine similarity, and $\tau$ is the temperature parameter.

To further enhance CLIP's performance on the downstream tasks, CoOp \cite{zhou2022learning} replaces hand-crafted prompts with a set of learnable textual soft prompts $\mathbf{P}=\{[\mathbf{P}]_l\}_{l=1}^L$, where each $[\mathbf{P}]_l$ denotes a learnable prompt vector. The resulting text prompt is then expressed by 
\begin{equation}
    \mathbf{t}(\mathbf{P}) = [\mathbf{P}]_1[\mathbf{P}]_2\ldots[\mathbf{P}]_L[\mathbf{CLS}].
\end{equation}
Accordingly, the probability of classifying image $\mathbf{x}$ as class $y_i$ is computed as:
\begin{equation}\label{probability}
    p(y_i|\mathbf{x}) = \frac{\exp(\langle \mathbf{z}_v, f_t(\mathbf{t}_{y_i}(\mathbf{P})) \rangle/\tau)}{\sum_{c=1}^{\vert C \vert} \exp (\langle\mathbf{z}_v, f_t(\mathbf{t}_c(\mathbf{P})) \rangle/\tau)},
\end{equation}
where $\mathbf{t}_c(\mathbf{P})$ denotes the learnable prompt corresponding to class $c$. Given the predicted probability distribution in Eq. (\ref{probability}), the cross-entropy (CE) loss is formulated as: 
\begin{equation}
    \mathcal{L}_{CE}(\mathbf{x}) = - \sum_{c=1}^{\vert C \vert} y_c \log p(c|\mathbf{x}),
\end{equation}
where $y_c$ is the binary ground-truth label (equal to 1 if $\mathbf{x}$ belongs to class $c$, and 0 otherwise).

\noindent\textbf{Risk definition.} To analyze our framework from the perspective of statistical learning theory, let $q_\phi(\mathbf{P}|\mathbf{x})$ denote the probabilistic mapping, parameterized by $\phi\in\Phi$, from the input $\mathbf{x}$ to the latent prompt $\mathbf{P} \in \mathcal{P}$, where $\mathcal{P}$ is the continuous prompt space and $\Phi$ is the parameter space of the encoder. Correspondingly, let $q_\theta(y|\mathbf{P})$ denote the decoder, parameterized by $\theta\in \Theta$, that projects the learned prompt to the class label, which mathematically abstracts the computation of Eq. (\ref{probability}). Here, $\Theta$ denotes the parameter space of the decoder.  After training on tasks $1,\ldots,t$, we denote the encoder parameters by $\phi^t$ and the decoder parameters by $\theta^t$. The overall predictive distribution of classifying $\mathbf{x}$ as class $y$ is defined as $P_{\phi, \theta}(y|\mathbf{x}) := \int_{\mathcal{P}} q_\theta(y|\mathbf{P}) q_\phi(\mathbf{P}|\mathbf{x}) d\mathbf{P}$. Given a loss $\ell: \Delta(\mathcal{C}) \times \mathcal{C} \rightarrow \mathbb{R}_{+}$, the population and empirical risks on task $t$ are defined as 
\begin{equation*}
\begin{split}
     \mathcal{R}_t(\phi,\theta)&:= \mathbb{E}_{(\mathbf{x},y)\sim \mathcal{D}_t}[\ell(P_{\phi, \theta}(\cdot|\mathbf{x}),y)], \\
     \quad \widehat{\mathcal{R}}_t(\phi,\theta)&:= \frac{1}{n_t}\sum_{i=1}^{n_t}[\ell(P_{\phi, \theta}(\cdot|\mathbf{x}_i),y_i)].
\end{split}
\end{equation*}
The forgetting on a previous task $t'<t$ after training task $t$ is then formulated by 
\begin{equation}
  F_{t'}(t) := \mathcal{R}_{t'}(\phi^t, \theta^t) - \mathcal{R}_{t'}(\phi^{t'}, \theta^{t'}).
\end{equation}
This measures the degree of performance degradation on task $t'$ after sequentially training on $t$ tasks. Further, let $\mathcal{Q} = \{ Q \mid Q(\cdot|\mathbf{x}) \in \Delta(\mathcal{C}), \forall \mathbf{x} \in \mathcal{X} \}$ denote the set of all measurable functions mapping the input space $\mathcal{X}$ to the space of probability distributions over $\mathcal{C}$. We define the Bayes risk on task $t$ as $\mathcal{R}_t^* := \inf_{Q \in \mathcal{Q}} \mathbb{E}_{(\mathbf{x},y)\sim \mathcal{D}_t} [\ell(Q(\cdot|\mathbf{x}), y)]$, which quantifies the minimum expected risk achievable by an arbitrary predictive function over the true data distribution $\mathcal{D}_t$. Throughout our theoretical analysis, we focus on the overall excess risk across $T$ tasks, defined as:
\begin{align}
    &\frac{1}{T}\sum_{t=1}^{T} \left( \mathcal{R}_t(\phi^T, \theta^T) - \mathcal{R}_t^* \right) \nonumber\\
  = & \underbrace{\frac{1}{T}\sum_{t=1}^{T} \left( \widehat{\mathcal{R}}_t(\phi^t, \theta^t) - \mathcal{R}_t^* \right)}_{\textit{Empirical Suboptimality Gap}} + \underbrace{\frac{1}{T}\sum_{t=1}^{T} \left[ \mathcal{R}_t(\phi^t, \theta^t) - \widehat{\mathcal{R}}_t(\phi^t, \theta^t) \right]}_{\textit{ Generalization Gap}} \nonumber\\
  &+ \underbrace{\frac{1}{T}\sum_{t=1}^{T} F_t(T)}_{\textit{ Forgetting}}, \label{decomposition}
\end{align}
which quantifies the discrepancy between the population risk of the learned model and that of the optimal functions in the hypothesis space $\mathcal{Q}$. Here, the excess risk can be decomposed into three terms, where the \textit{Empirical Suboptimality Gap} term measures how closely the empirical performance over $T$ tasks aligns with the theoretical best performance, the \textit{Generalization Gap} characterizes the model's ability to transfer knowledge learned on training data to unseen data, and the \textit{Forgetting} reflects the magnitude of catastrophic forgetting incurred during incremental training.

\begin{figure*}[tb]
  \centering
  \includegraphics[width=\linewidth]{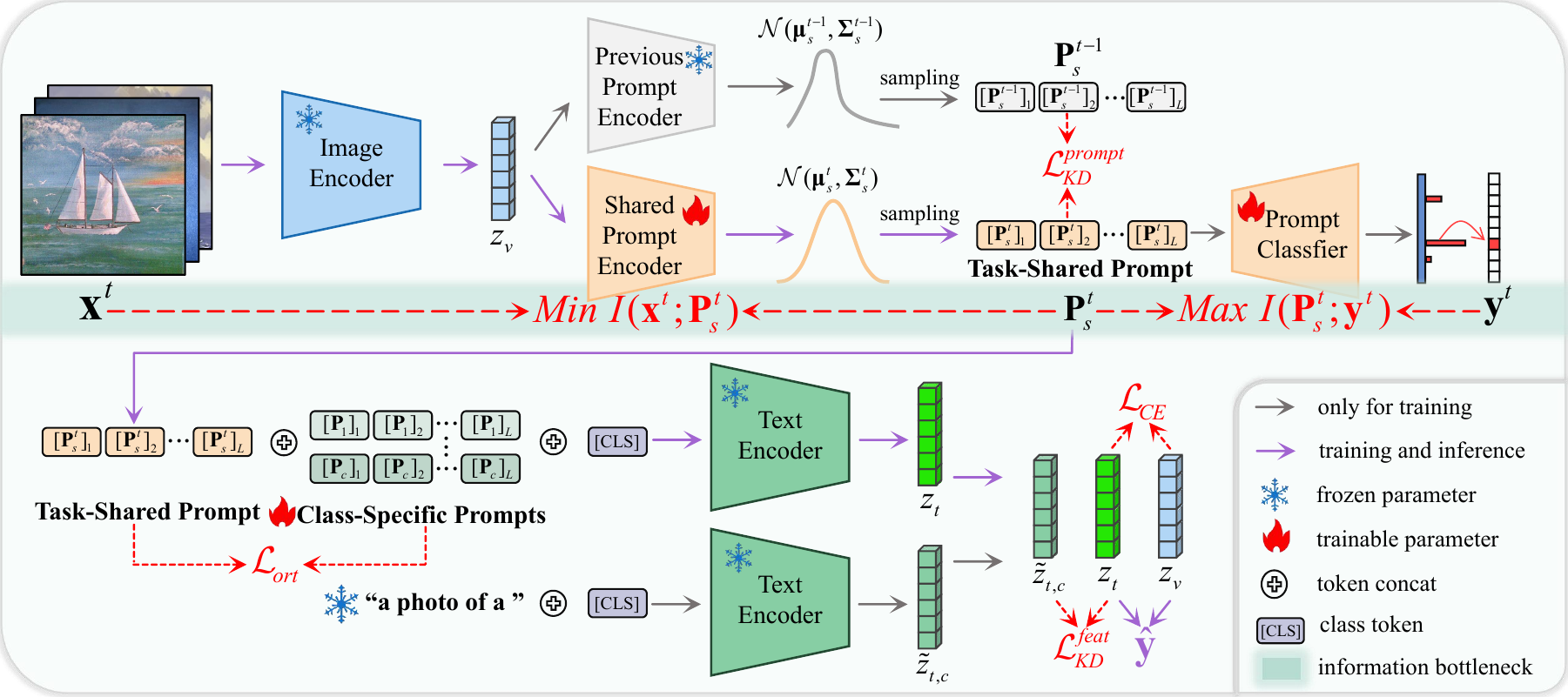}
  \caption{Overall framework of the proposed DPT4CL. The IB loss $\mathcal{L}_{IB}$ is implemented through a max--min MI objective, which learns a task-shared prompt distribution to facilitate knowledge transfer across sequential tasks. The distributional distillation loss $\mathcal{L}_{KD}$ maintains semantic consistency to alleviate classifier bias, while the orthogonalization loss $\mathcal{L}_{ort}$ maximally disentangles class-specific prompts from the task-shared prompt to enhance inter-class separability. The classification loss $\mathcal{L}_{CE}$ ensures semantic discriminability. During inference, the prediction $\hat{y}$ is obtained by computing the cosine similarity between the visual feature $\mathbf{z}_v$ and the textual feature $\mathbf{z}_t$ generated from the concatenated prompts.}
  \label{fig:framework}
\end{figure*}

\subsection{Framework of DPT4CL}
Following the motivation discussed above, we present the concrete decoupling implementation of DPT4CL. Specifically, let $\mathbf{P}_s$ and $\mathbf{P}_c$ denote the task-shared prompt and class-specific prompts, respectively, where $c \in {1, \dots, |C|}$. These prompts are constrained to satisfy the following conditional independence with orthogonal constraints:
\begin{proposition}\label{proposition1}
  A prompt $\mathbf{P}=(\mathbf{P}_s,\mathbf{P}_1,\ldots,\mathbf{P}_{|C|})$ is said to be maximally disentangled if the class-specific components $\mathbf{P}_c$, for $c=1,\ldots,|C|$ are conditionally independent given the task-shared consensus $\mathbf{P}_s$. This implies a joint distribution $p(\mathbf{P}) = p(\mathbf{P}_s) \prod_{c=1}^{|C|} p(\mathbf{P}_c|\mathbf{P}_s)$, which is geometrically enforced by the constraint that the consensus subspace $\mathcal{S}_{s}$ and all class-specific subspaces $\{\mathcal{S}_{c}\}_{c=1}^{|C|}$ are mutually orthogonal.
\end{proposition}

To effectively capture the consensus component $\mathbf{P}_s$ across different tasks, we employ the IB principle \cite{tishby2000information} to distill task-invariant consensus into the task-shared prompt distribution $\mathbf{p}$ by maximizing the MI with the class label $\mathbf{y}$ while minimizing the MI with the input image $\mathbf{x}$. This objective ensures that the learned distribution $\mathbf{p}$ extracts the minimal sufficient statistics necessary for robust cross-task knowledge transfer, which is defined as:
\begin{equation}\label{IB}
    \max I(\mathbf{p};\mathbf{y}) - \beta \cdot I(\mathbf{p}; \mathbf{x}),
\end{equation}
where $\beta\geq 0$ is the Lagrange multiplier and $I(\cdot;\cdot)$ denotes MI. It is noteworthy that directly estimating the MI terms in Eq. (\ref{IB}) is computationally intractable in high-dimensional spaces. To this end, we adopt the variational formulation introduced in VIB \cite{alemi2016deep} to derive a lower bound for the objective in Eq. (\ref{IB}). For brevity, we place the derivation in the Supplementary Material and only show the result of derivation:
\begin{align} \label{lower}
    I(\mathbf{p};\mathbf{y}) \geq \mathbb{E}_{p(\mathbf{p},\mathbf{y})} [\log q_\theta (\mathbf{y} |\mathbf{p})] + H(\mathbf{y}),
\end{align}
where $p(\mathbf{p},\mathbf{y})$ denotes the joint distribution of $\mathbf{p}$ and $\mathbf{y}$, and $q_\theta(\mathbf{y}|\mathbf{p})$ is a variational approximation to the conditional distribution $p(\mathbf{y}|\mathbf{p})$, which is parameterized by the prompt classifier illustrated in Fig. \ref{fig:framework}, and $H(\mathbf{y})$ denotes the entropy of the class label. Since $H(\mathbf{y})$ is independent of the parameter optimization and can thus be ignored, we can instead maximize the lower bound in Eq. (\ref{lower}) as a surrogate for directly optimizing $I(\mathbf{p};\mathbf{y})$. For the second term in Eq. (\ref{IB}), we also place the detailed derivation in the Supplementary Material and only present the result of derivation:
\begin{align} \label{upper}
    I(\mathbf{p}; \mathbf{x}) \leq \mathbb{E}_{p(\mathbf{x})} [\mathrm{KL}(q_\phi(\mathbf{p}|\mathbf{x}) \| q(\mathbf{p}))],
\end{align}
where $\mathrm{KL}(\cdot \| \cdot)$ denotes the KL divergence, while $q_\phi(\mathbf{p}|\mathbf{x})$ and $q(\mathbf{p})$ are variational approximations to the conditional distribution $p(\mathbf{p}|\mathbf{x})$ and the marginal distribution $p(\mathbf{p})$, respectively. Here, we utilize a standard normal Gaussian distribution for the approximation, namely, $q(\mathbf{p})\sim \mathcal{N}(0,\mathbf{I})$, and assume that $q_\phi(\mathbf{p}|\mathbf{x})\sim \mathcal{N}(\boldsymbol{\mu}_s,\mathbf{\Sigma}_s)$, where the mean $\boldsymbol{\mu}_s$ and covariance $\mathbf{\Sigma}_s$ are parameterized and learned via a task-shared prompt encoder \cite{alemi2016deep,wang2019deep} as shown in Fig. \ref{fig:framework}. 

Combining the above estimations, we can obtain a lower bound of the objective function in Eq. (\ref{IB}):
\begin{align}
    \mathcal{L}_{IB}(\mathbf{x}, \mathbf{p}) 
    &= \mathbb{E}_{p(\mathbf{x}, \mathbf{y})} \Big[ -\mathbb{E}_{q_\phi(\mathbf{p}|\mathbf{x})}[\log q_\theta (\mathbf{y} |\mathbf{p})] \nonumber \\
    &\quad + \beta \mathrm{KL}(q_\phi(\mathbf{p}|\mathbf{x}) \| q(\mathbf{p})) \Big] \nonumber \\
    &= \mathbb{E}_{p(\mathbf{x}, \mathbf{y})} \Big[ -\mathbb{E}_{q_\phi(\mathbf{p}|\mathbf{x})}[\log q_\theta (\mathbf{y} |\mathbf{p})] \nonumber \\
    &\quad + \beta \mathrm{KL}(\mathcal{N}(\boldsymbol{\mu}_s,\mathbf{\Sigma}_s) \| \mathcal{N}(0,\mathbf{I})) \Big], \label{objective}
\end{align}
which serves as a tractable surrogate objective for learning the prompt distribution. Notably, the first term in Eq. (\ref{objective}) corresponds to the prompt classification loss, while the second term acts as a regularizer, encouraging the learned distribution to approximate the prior.

To mitigate classifier bias induced by independently optimizing the distribution parameters $(\boldsymbol{\mu}_s, \boldsymbol{\Sigma}_s)$ across tasks in a CoOp-like manner, we utilize a distribution distillation term that leverages the task-shared prompt to minimize the KL divergence between the current variational posterior and that from the previous task, thereby enforcing a unified task-shared prompt distribution. Furthermore, we introduce a language-aware distillation loss to align the text features derived from the learned prompts with the original feature space. This effectively mitigates cross-modal deviation \cite{jha2024clap4clip} between the learned text representations and the frozen visual features, thereby preserving CLIP's inherent representation power and preventing feature-level forgetting.

Specifically, consider the current learned distribution and the reparameterized task-shared prompt $\mathcal{N}(\boldsymbol{\mu}^t_s, \boldsymbol{\Sigma}^t_s)$, $\mathbf{P}^t_s = \boldsymbol{\mu}^t_s + \boldsymbol{\Sigma}^t_s \odot \boldsymbol{\epsilon}$ along with their corresponding distribution and the reparameterized prompt from the end of the previous task $\mathcal{N}(\boldsymbol{\mu}^{t-1}_s, \boldsymbol{\Sigma}^{t-1}_s)$, $\mathbf{P}_s^{t-1} = \boldsymbol{\mu}^{t-1}_s + \boldsymbol{\Sigma}^{t-1}_s \odot \boldsymbol{\epsilon}$, where $\boldsymbol{\epsilon} \sim \mathcal{N}(0,\mathbf{I})$. The distillation loss can be formulated as:
\begin{equation}\label{distillation_loss}
    \mathcal{L}_{KD}(\mathbf{p}) =  \underbrace{\mathrm{KL}\Big(\mathbf{P}_s^{t}  \Big\| \mathbf{P}_s^{t-1} \Big)}_{\mathcal{L}_{KD}^{prompt}} + \underbrace{\sum_{c=1}^{\vert C \vert} \mathrm{KL}\Big(f_t\big(\mathbf{t}_c(\mathbf{P}^{t}_s,\mathbf{P}^{t}_c )\big)  \Big\| \tilde{\mathbf{z}}_{t,c} \Big)}_{\mathcal{L}_{KD}^{feat}},
\end{equation}
where $f_t\big(\mathbf{t}_c(\mathbf{P}^{t}_s,\mathbf{P}^{t}_c )\big)$ is the projected text representation of the $c$-th class and $\tilde{\mathbf{z}}_{t,c}$ denotes the corresponding hand-crafted prompt-based representation. Here, prompt-level distribution distillation $\mathcal{L}_{KD}^{prompt}$ penalizes the divergence of the prompt distribution, and feature-level language-aware distillation $\mathcal{L}_{KD}^{feat}$ anchors the encoded text features to the original semantic space. Together, these two terms maintain semantic consistency across tasks and between modalities, thereby alleviating the classifier bias caused by CoOp-like prompt optimization.

Regarding the class-specific prompts $\mathbf{P}_c$ for $c \in \{1, \ldots, |C|\}$, we employ an orthogonalization objective $\mathcal{L}_{ort}$ to ensure that each prompt captures distinct, class-exclusive information while remaining strictly decoupled from the consensus knowledge preserved in the task-shared prompt $\mathbf{P}_s$. Specifically, we concatenate all prompt vectors into a matrix $\mathcal{P} = \{\mathbf{P}_s, \mathbf{P}_1, \ldots, \mathbf{P}_{|C|}\}$, and the objective is formulated as
\begin{equation}\label{L_ort}
  \mathcal{L}_{ort} = \sum_{\mathbf{P}_i, \mathbf{P}_j \in \mathcal{P}, i \neq j} \left( \frac{\langle \mathbf{P}_i, \mathbf{P}_j \rangle}{\|\mathbf{P}_i\|_2 \|\mathbf{P}_j\|_2} \right)^2.
\end{equation}
$\mathcal{L}_{ort}$ serves as the structural constraint to ensure that the consensus knowledge and class-specific information encoded within the prompts are mutually orthogonal, as required by Proposition \ref{proposition1}, thereby promoting the learning of prompts that are both sufficient and maximally disentangled.

Consequently, by substituting the learnable prompt $\mathbf{P}$ in Eq. (\ref{probability}) with the concatenation of the task-shared prompt $\mathbf{P}_s$ and the class-specific prompt $\mathbf{P}_c$, the overall training objective for DPT4CL is formulated as:
\begin{equation}
\mathcal{L}_{\text{total}}^{\text{DPT4CL}} = \mathcal{L}_{CE}(\mathbf{x}) + \alpha \mathcal{L}_{IB}(\mathbf{x}, \mathbf{p}) + \lambda \mathcal{L}_{KD}(\mathbf{p}) + \eta \mathcal{L}_{ort},
\label{DPT4CL}
\end{equation}
where $\alpha$, $\lambda$, and $\eta$ are regularization coefficients.

During inference, a test image $\mathbf{x}$ is first fed into the frozen image encoder to obtain its visual representation $\mathbf{z}_v=f_v(\mathbf{x})$. Meanwhile, the shared prompt encoder samples the task-shared prompt $\mathbf{P}_s$ conditioned on $\mathbf{z}_v$. For each observed class $c \in \mathcal{C}$, the sampled task-shared prompt $\mathbf{P}_s$, the corresponding class-specific prompt $\mathbf{P}_c$, and the class token $[\text{CLS}]_c$ are concatenated and then fed into the text encoder to obtain the textual representation 
$\mathbf{z}_{t,c} = f_t\big(t_c(\mathbf{P}_s,\mathbf{P}_c)\big)$, where $t_c(\mathbf{P}_s,\mathbf{P}_c)\!=\!\operatorname{concat}(\mathbf{P}_s;\mathbf{P}_c;[\mathrm{CLS}]_c)$ denotes the textual prompt constructed for class $c$.  The probability that $\mathbf{x}$ belongs to the $i$-th class is computed as
\begin{equation}
p(y_i|\mathbf{x}) =
\frac{
\exp\left(\langle \mathbf{z}_v, \mathbf{z}_{t,i} \rangle / \tau \right)
}{\sum_{c=1}^{\vert C \vert}
\exp\left(\langle \mathbf{z}_v, \mathbf{z}
_{t,c} \rangle / \tau \right)
}.
\label{infer_prob}
\end{equation}
The final prediction is then given by
\begin{equation} \hat{y} =
\operatorname*{argmax}_{c \in \mathcal{C}} p(y_c|\mathbf{x}).
\label{infer_pred}
\end{equation}

\subsection{Excess Risk Analysis}
In this section, we establish a unified excess risk bound by separately controlling the \textit{Empirical Suboptimality Gap}, \textit{Generalization Gap}, and \textit{Forgetting} terms in Eq. (\ref{decomposition}), thereby providing a rigorous theoretical guarantee for the proposed framework.

For each task $t$, let $\mathcal{R}^{\phi^*,\theta^*}_{t} := \mathcal{R}_t(\phi^{t*}, \theta^{t*})$ denote the task-specific optimal population risk over the parameterized hypothesis spaces $\Phi$ and $\Theta$, where the optimal parameters are defined as $(\phi^{t*}, \theta^{t*}) = \arg\min_{\phi \in \Phi, \theta \in \Theta} \mathbb{E}_{(\mathbf{x},y) \sim \mathcal{D}_t} [\ell(P_{\phi,\theta}(\cdot|\mathbf{x}), y)]$. Similarly, let $q_\beta^*(p|\mathbf{x})$ denote the minimal sufficient statistic encoder obtained at the stationary point of the IB Lagrangian (\ref{IB}). Its corresponding Bayes risk under optimal decoding is denoted as $\mathcal{R}_t^{q_\beta^*}$. Based on the above definition, we have the following upper bound on the \textit{Empirical Suboptimality Gap}:
\begin{theorem}[Empirical Suboptimality Gap]\label{Theorem1}
   Assume that the loss function $\ell(\cdot,\cdot)\in[0,1]$ is $L$-Lipschitz continuous. For any $\delta\in(0,1)$, with probability at least $1-\delta$ over the datasets $\{D_t\}_{t=1}^T$, we have 
    \begin{equation}
    \begin{split}
        &\frac{1}{T}\sum_{t=1}^{T} \left( \widehat{\mathcal{R}}_t(\phi^t, \theta^t) - \mathcal{R}_t^* \right) \\
        &\le \frac{1}{T}\sum_{t=1}^{T} \left[ \widehat{\mathcal{E}}_{\text{opt}}^{(t)} + \epsilon_{IB}(\beta) + \Delta_{\text{arch}}^{(t)} + \sqrt{\frac{\log(T/\delta)}{2n_t}} \right],
    \end{split}
    \end{equation}
where $\widehat{\mathcal{E}}_{\text{opt}}^{(t)} := \widehat{\mathcal{R}}_t(\phi^t, \theta^t) - \widehat{\mathcal{R}}_t(\phi^{t*}, \theta^{t*})$, $\epsilon_{IB}(\beta) := \mathcal{R}^{\phi^*,\theta^*}_{t} - \mathcal{R}_t^{q_\beta^*}$, and $\Delta_{\text{arch}}^{(t)} := \mathcal{R}_t^{q_\beta^*} - \mathcal{R}_t^*$.
\end{theorem}

\begin{remark}
  Theorem \ref{Theorem1} establishes the upper bound on the \textit{Empirical Suboptimality Gap}, which is governed by the empirical optimization error $\widehat{\mathcal{E}}_{\text{opt}}^{(t)}$, the IB representation error $\epsilon_{IB}(\beta)$, and the irreducible architectural bias $\Delta_{\text{arch}}^{(t)}$. These components provide a theoretical justification for the effectiveness of the proposed objectives (\ref{DPT4CL}) in improving performance. Specifically, $\widehat{\mathcal{E}}_{\text{opt}}^{(t)}$ quantifies the discrepancy between the learned model and the ideal hypothesis on the training set, capturing the optimization error inherent in the empirical training process. Given that the risk $\widehat{\mathcal{R}}_t(\phi^{t*}, \theta^{t*})$ w.r.t the global population minimizer $\phi^{t*}, \theta^{t*}$ typically approaches zero, one can directly control $\widehat{\mathcal{E}}_{\text{opt}}^{(t)}$ by minimizing the empirical risk $\widehat{\mathcal{R}}_t(\phi^t, \theta^t)$, which is exactly achieved by the cross-entropy objective $\mathcal{L}_{CE}$ in our framework. 
  
Furthermore, $\epsilon_{IB}(\beta)$ characterizes representational discrepancy between the learned compressed latent representations and the ideal IB sufficient statistic $q_\beta^*$. A large $\epsilon_{IB}(\beta)$ indicates the retention of task-irrelevant redundancy or the loss of task-discriminative information. This discrepancy is effectively mitigated by minimizing a variational upper bound of the IB Lagrangian, defined as $\mathcal{L}_{IB}$ in Eq. (\ref{objective}), which enforces a trade-off between compression and predictive precision. 

The other factor $\Delta_{\text{arch}}^{(t)}$ measures the inherent structural bias between the IB-optimal trade-off and the non-parametric Bayes risk. This gap fundamentally arises from the inherent limitations of the parameterized prompt space in capturing the maximal sufficient statistic of the underlying data distribution. Specifically, by Fano's inequality \cite{cover1999elements,nikolopoulos2025minimum}, the classification error is lower-bounded by the conditional entropy $H(y|\mathbf{P})$, that is, the risk lower bound tightens as $H(y|\mathbf{P})$ decreases. Given $I(\mathbf{P}; y) = H(y) - H(y|\mathbf{P})$, maximizing the mutual information $I(\mathbf{P}; y)$ can effectively reduce $H(y|\mathbf{P})$, thereby tightening the risk lower bound. Hence, it is necessary to apply orthogonal constraints $\mathcal{L}_{ort}$ in (\ref{L_ort}) to capture sufficient, effective discriminative information while eliminating inter-prompt redundancy, mitigating the architectural approximation error $\Delta_{\text{arch}}^{(t)}$ relative to the non-parametric Bayes limit. 
\end{remark}

In the following theorem, we establish an upper bound on the \textit{Generalization Gap} defined in Eq. (\ref{decomposition}) by leveraging the PAC-Bayesian framework.

\begin{theorem}[Generalization Gap]\label{Theorem2}
Assume that the loss function $\ell(\cdot,\cdot)\in[0,1]$ is $L$-Lipschitz continuous with respect to the model parameters. Let $(\phi^t, \theta^t)$ be the parameters learned at task $t \in [T]$. Let $Q^t$ denote a data-dependent distribution over the parameter space centered at $(\phi^t, \theta^t)$ with an expected perturbation radius $\rho$, such that $\mathbb{E}_{(\phi,\theta) \sim Q^t} \Vert (\phi,\theta) - (\phi^t, \theta^t) \Vert \le \rho$. Let $Q^{t-1}$ denote the corresponding prior distribution inherited from task $t-1$, where $Q^0$ is a data-independent initial prior. For any $\delta \in (0,1)$, with probability at least $1-\delta$ over the joint generation of datasets $\{D_t\}_{t=1}^T$, we have 
\begin{align}
     & \frac{1}{T}\sum_{t=1}^{T} \left[ \mathcal{R}_t(\phi^t, \theta^t) - \widehat{\mathcal{R}}_t(\phi^t, \theta^t) \right] \nonumber\\
    \le & \frac{1}{T}\sum_{t=1}^{T} \left( \sqrt{\frac{\mathrm{KL}(Q^t \Vert Q^{t-1}) + \log(2T\sqrt{n_t}/\delta)}{2n_t}} + 2L\rho \right),
\end{align}
where $\mathrm{KL}(\cdot \Vert \cdot)$ is the Kullback-Leibler divergence.
\end{theorem}

\begin{remark}
  Theorem \ref{Theorem2} suggests that the \textit{Generalization Gap} is controlled by the PAC-Bayes complexity term, represented by the Kullback-Leibler divergence $\mathrm{KL}(Q^t \Vert Q^{t-1})$ between the consecutive parameter distributions. Since direct optimization in the high-dimensional space $\Phi \times \Theta$ is computationally intractable, we introduce distribution distillation as a tractable surrogate, which projects the parameter distributions onto the functional prompt-encoder space. This strategy effectively reduces the intractable parameter-level divergence into a computable divergence within the latent prompt space. In particular, for a given input $\mathbf{x}$, the divergence is evaluated between the reparameterized prompt distributions $\mathcal{N}(\boldsymbol{\mu}_s^t, \boldsymbol{\Sigma}_s^t)$ and $\mathcal{N}(\boldsymbol{\mu}_s^{t-1}, \boldsymbol{\Sigma}_s^{t-1})$, having 
$\mathrm{KL}(Q^t \parallel Q^{t-1}) \lesssim \mathbb{E}_{\mathbf{x} \sim \mathcal{D}_t} \left[ \mathrm{KL} \big( q_{\phi^t}(\mathbf{P}|\mathbf{x}) \parallel q_{\phi^{t-1}}(\mathbf{P}|\mathbf{x}) \big) \right],$
which is formulated as the prompt-level regularization in the distillation loss $\mathcal{L}_{KD}$ in Eq. (\ref{distillation_loss}). Consequently, minimizing this regularizer can tighten the generalization bound, intrinsically mitigating catastrophic forgetting while facilitating task-shared knowledge transfer.
\end{remark}

The following theorem provides the information-theoretic upper bound for the \textit{Forgetting} term formulated in Eq. (\ref{decomposition}).

\begin{theorem}[Average Forgetting Bound]\label{Theorem3}
Assume that the loss function $\ell(\cdot,\cdot)\in[0,1]$ is $L$-Lipschitz continuous. Let $\phi^T$ and $\phi^t$ denote the encoder parameters learned at the task $T$ and a previous task $t$. Under a task-specific decoder architecture where the decoder parameters $\theta^t$ assigned to task $t$ remain fixed during the training of subsequent tasks, the average catastrophic forgetting across all $T$ tasks is upper-bounded by:
\begin{equation}
   \frac{1}{T}\sum_{t=1}^{T} F_t(T) \le  \frac{L}{T}\sum_{t=1}^{T-1} \sqrt{\frac{1}{2} \mathbb{E}_{\mathbf{x} \sim \mathcal{D}_t} \left[ \mathrm{KL} \big( q_{\phi^T}(\cdot|\mathbf{x}) \parallel q_{\phi^t}(\cdot|\mathbf{x}) \big) \right]}.
\end{equation}
\end{theorem}

\begin{remark}
Notably, Theorem \ref{Theorem3} reveals that the average forgetting is governed by the exact same KL divergence between the latent prompt distributions, i.e., $\mathbb{E}_{\mathbf{x} \sim \mathcal{D}_t} \left[ \mathrm{KL} \big( q_{\phi^T}(\cdot|\mathbf{x}) \parallel q_{\phi^t}(\cdot|\mathbf{x}) \big) \right]$, as the PAC-Bayesian complexity in Theorem \ref{Theorem2}. Since directly optimizing the parameter drift in the high-dimensional space $\Phi$ is computationally intractable, the prompt-level distillation term defined in Eq. (\ref{distillation_loss}) serves as a sufficient and tractable surrogate to effectively alleviate catastrophic forgetting. Additionally, to prevent these stable prompts from exhibiting cross-modal deviations, we empirically introduce the feature-level distillation term to constrain the learned prompts within the pre-trained semantic space.  Consequently, minimizing the unified distillation loss $\mathcal{L}_{KD}$ in Eq. (\ref{distillation_loss}) both theoretically and empirically guarantees the ability to mitigate catastrophic forgetting while preserving the zero-shot recognition ability of pre-trained models.
\end{remark}

Substituting Theorems \ref{Theorem1}, \ref{Theorem2}, and \ref{Theorem3} into Eq. (\ref{decomposition}), we obtain the desired unified excess risk bound, which is omitted due to space constraints.

\section{Experiments}
\subsection{Implementation Details}
\subsubsection{Datasets}
We evaluate our methods on four widely used CL benchmarks including CIFAR-100 \cite{krizhevsky2009cifar}, ImageNet-R \cite{hendrycks2021many}, CUB-200 \cite{wah2011cub200}, and UCF-101 \cite{soomro2012ucf101}. Following \cite{zhou2025external}, we select $100$ classes from UCF-101 for evaluation. Detailed information and statistics of the datasets are provided in the Supplementary Material. For all datasets, we adopt a class-incremental setup with 10 tasks and report the average results over three random seeds. To ensure a fair comparison with prior work \cite{chen2025preserving,zhou2025external}, we additionally report the performance of all methods under the random seed 1993 in the Supplementary Material.

\subsubsection{Evaluation metric} Overall average accuracy ($\textrm{Avg.}$), final accuracy ($\textrm{Last}$), and average forgetting measure ($\textrm{FM}$) are utilized to evaluate our model. Specifically, we adopt the $\textrm{Avg.}$ and $\textrm{Last}$ to quantify the generalization performance across all $T$ tasks, defined as $\textrm{Avg.} = \frac{1}{T} \sum_{t=1}^{T} A_t$ and $\textrm{Last} = A_T$, respectively, where $A_t$ is the average accuracy over observed classes $\bigcup_{i=1}^{t}C_{i}$ of $t$ tasks. Furthermore, we utilize the $\textrm{FM}$ to quantify the extent of catastrophic forgetting on previously learned tasks, defined as $\textrm{FM} = \frac{1}{T-1} \sum_{i=1}^{T-1} \left(a_i^* - a_{i,T}\right)$, where $a_i^*$ denotes the maximum accuracy on task $i$, and $a_{i,T}$ denotes the test accuracy of task $i$ after training on the final task $T$. A lower $\textrm{FM}$ indicates better retention of knowledge.

\subsubsection{Baselines and compared methods}
As summarized in Table~\ref{tab:main_results}, we compare our proposed DPT4CL with a wide range of CL methods, which can be grouped into three categories: (1) \textit{Traditional CL methods.} We adopt LwF \cite{li2018learning}, iCaRL \cite{rebuffi2017icarl}, DER \cite{yan2021der}, and TagFex \cite{zheng2025task} as representative traditional CL competitors. (2) \textit{Prompt-based methods.} Continual-CLIP \cite{thengane2022clip} serves as the zero-shot recognition lower bound for CLIP-based methods. CoOp \cite{zhou2022learning} learns only task-shared prompts in the textual branch of CLIP. AttriCLIP~\cite{wang2023attriclip}, L2P~\cite{wang2022learning}, DualPrompt~\cite{wang2022dualprompt}, and CODA-Prompt~\cite{smith2023coda} are included as representative methods that learn task-specific prompt pools. (3) \textit{CLIP-based SOTA methods.} We compare with state-of-the-art CLIP-based methods, including CLAP4CLIP~\cite{jha2024clap4clip}, ENGINE~\cite{zhou2025external}, and BOFA~\cite{li2026bofa}. Additionally, to estimate the performance upper bound of our method, we evaluate DPT4CL under a joint-training setting, where all tasks are trained simultaneously and catastrophic forgetting is avoided.

\subsubsection{Training details}
We implement our method using PyTorch and conduct all experiments on an NVIDIA A800 GPU. To ensure a fair comparison, all compared methods are re-implemented with the same OpenAI-pre-trained CLIP ViT-B/16 backbone. For rehearsal-based methods, we use a fixed memory buffer of 2,000 samples, which are equally partitioned among all encountered classes. Optimization is performed using SGD, with momentum and weight decay set to 0.9 and 0.05, respectively. The learning rate is initialized to 0.05 and decayed by cosine annealing. For each incremental stage, the model is trained for 10 epochs with a batch size of 32. The reparameterization \cite{alemi2016deep,wang2019deep} in Eq.~(\ref{objective}) is implemented by a task-shared prompt encoder parameterized by a two-layer MLP, followed by a prompt classifier implemented by another two-layer MLP decoder, which is reinitialized at each new stage. The length $L$ of both the task-shared prompt $\mathbf{P}_s$ and the class-specific prompt $\mathbf{P}_c$ is fixed to 2. The dimension of each prompt vector is set to match the feature dimension of the pre-trained CLIP encoders for seamless multimodal interaction. Through grid search, the hyperparameters $\alpha$, $\beta$, $\lambda$, and $\eta$ in Eqs.~(\ref{objective}) and~(\ref{DPT4CL}) are set to 0.5, $2\times10^{-3}$, 1, and 10, respectively.

\begin{table*}[t]
\centering
\caption{Comparison of Avg., Last, and FM (\%) with baselines and state-of-the-art methods on CIFAR-100, ImageNet-R, CUB-200, and UCF-101 under a 10-task class-incremental learning setting. Best results are marked in \textbf{bold}. Second-best results are \underline{underscored}.}
\label{tab:main_results}
\small
\setlength{\tabcolsep}{1.6pt}
\renewcommand{\arraystretch}{1.12}
\resizebox{\textwidth}{!}{
\begin{tabular}{@{}l c c c c c c c c c c c c c@{}}
    \toprule
    \multirow{2}{*}{Method} & \multirow{2}{*}{Memory} & \multicolumn{3}{c}{CIFAR-100} & \multicolumn{3}{c}{ImageNet-R} & \multicolumn{3}{c}{CUB-200} & \multicolumn{3}{c}{UCF-101} \\
    \cmidrule(lr){3-5} \cmidrule(lr){6-8} \cmidrule(lr){9-11} \cmidrule(lr){12-14}
    & & Avg. $\uparrow$ & Last $\uparrow$ & FM $\downarrow$ & Avg. $\uparrow$ & Last $\uparrow$ & FM $\downarrow$ & Avg. $\uparrow$ & Last $\uparrow$ & FM $\downarrow$ & Avg. $\uparrow$ & Last $\uparrow$ & FM $\downarrow$ \\
    \midrule
    LwF~\cite{li2018learning} & 2000 & 40.90$_{\pm3.69}$ & 28.75$_{\pm4.27}$ & 32.93$_{\pm0.96}$ & 26.92$_{\pm2.87}$ & 19.02$_{\pm2.13}$ & 20.50$_{\pm1.18}$ & 20.15$_{\pm6.48}$ & 14.72$_{\pm3.30}$ & 21.22$_{\pm4.20}$ & 42.50$_{\pm6.19}$ & 24.55$_{\pm2.58}$ & 45.99$_{\pm2.01}$ \\
    iCaRL~\cite{rebuffi2017icarl} & 2000 & 45.51$_{\pm7.82}$ & 27.67$_{\pm9.71}$ & 54.43$_{\pm3.49}$ & 23.93$_{\pm13.26}$ & 11.17$_{\pm6.80}$ & 21.96$_{\pm8.80}$ & 44.84$_{\pm12.67}$ & 35.74$_{\pm11.28}$ & 18.79$_{\pm2.34}$ & 79.19$_{\pm8.79}$ & 69.16$_{\pm10.66}$ & 19.77$_{\pm6.56}$ \\
    DER~\cite{yan2021der} & 2000 & 40.07$_{\pm4.80}$ & 19.18$_{\pm3.33}$ & 69.63$_{\pm1.23}$ & 15.70$_{\pm8.58}$ & 3.54$_{\pm1.39}$ & 21.76$_{\pm9.82}$ & 20.55$_{\pm3.17}$ & 9.41$_{\pm1.36}$ & 27.04$_{\pm2.60}$ & 66.83$_{\pm5.61}$ & 37.11$_{\pm3.93}$ & 61.40$_{\pm1.42}$ \\
    TagFex~\cite{zheng2025task} & 2000 & 57.74$_{\pm8.24}$ & 49.11$_{\pm4.32}$ & 45.88$_{\pm0.80}$ & 13.03$_{\pm2.34}$ & 3.52$_{\pm0.59}$ & 19.73$_{\pm3.54}$ & 14.00$_{\pm4.37}$ & 5.20$_{\pm0.55}$ & 22.20$_{\pm4.10}$ & 63.95$_{\pm4.08}$ & 35.82$_{\pm3.04}$ & 62.14$_{\pm2.17}$ \\
    \midrule
    Continual-CLIP~\cite{thengane2022clip} & --- & 78.79$_{\pm0.32}$ & 68.37$_{\pm0.03}$ & 8.67$_{\pm1.01}$ & 79.74$_{\pm0.45}$ & 72.98$_{\pm0.00}$ & 9.02$_{\pm0.96}$ & 64.70$_{\pm1.32}$ & 53.29$_{\pm0.07}$ & 12.95$_{\pm2.96}$ & 74.92$_{\pm1.80}$ & 66.33$_{\pm0.04}$ & 9.59$_{\pm0.57}$ \\
    CoOp~\cite{zhou2022learning} & --- & 77.22$_{\pm0.39}$ & 64.89$_{\pm2.23}$ & 18.98$_{\pm2.30}$ & 81.83$_{\pm0.43}$ & 74.32$_{\pm1.73}$ & 9.98$_{\pm1.01}$ & 64.08$_{\pm1.02}$ & 47.57$_{\pm2.87}$ & 23.16$_{\pm2.96}$ & 76.07$_{\pm0.79}$ & 64.04$_{\pm0.34}$ & 27.48$_{\pm1.88}$ \\
    AttriCLIP~\cite{wang2023attriclip} & --- & 78.76$_{\pm0.13}$ & 66.34$_{\pm1.37}$ & 17.93$_{\pm1.16}$ & 82.90$_{\pm0.33}$ & 76.55$_{\pm0.68}$ & 8.31$_{\pm0.68}$ & 65.57$_{\pm1.13}$ & 50.33$_{\pm1.57}$ & 18.70$_{\pm0.37}$ & 79.99$_{\pm0.86}$ & 68.48$_{\pm2.14}$ & 20.62$_{\pm2.62}$ \\
    L2P~\cite{wang2022learning} & --- & 80.73$_{\pm1.13}$ & 68.07$_{\pm1.24}$ & 8.60$_{\pm2.06}$ & 76.83$_{\pm1.33}$ & 66.95$_{\pm2.23}$ & 13.38$_{\pm1.93}$ & 65.84$_{\pm1.57}$ & 52.52$_{\pm1.06}$ & 16.43$_{\pm3.58}$ & 84.52$_{\pm0.69}$ & 75.47$_{\pm0.16}$ & 14.71$_{\pm2.35}$ \\
    DualPrompt~\cite{wang2022dualprompt} & --- & 83.07$_{\pm1.01}$ & 73.04$_{\pm0.53}$ & \textbf{6.02$_{\pm0.94}$} & 82.53$_{\pm0.73}$ & 75.68$_{\pm0.74}$ & 9.40$_{\pm0.54}$ & 70.88$_{\pm1.84}$ & 58.59$_{\pm1.51}$ & 15.77$_{\pm1.50}$ & 91.32$_{\pm1.17}$ & 84.82$_{\pm0.91}$ & 7.28$_{\pm1.37}$ \\
    CODA-Prompt~\cite{smith2023coda} & --- & 83.48$_{\pm0.76}$ & 72.04$_{\pm1.43}$ & 16.81$_{\pm1.48}$ & 82.54$_{\pm0.21}$ & 74.47$_{\pm0.44}$ & 9.83$_{\pm0.94}$ & 72.04$_{\pm1.55}$ & 58.23$_{\pm1.00}$ & 19.86$_{\pm1.75}$ & 88.90$_{\pm1.19}$ & 80.69$_{\pm0.83}$ & 15.72$_{\pm0.94}$ \\
    \midrule
    CLAP4CLIP~\cite{jha2024clap4clip} & 2000 & \underline{83.85$_{\pm0.29}$} & 75.04$_{\pm0.43}$ & 9.04$_{\pm0.68}$ & 84.19$_{\pm0.38}$ & 78.57$_{\pm0.30}$ & \underline{6.86$_{\pm0.96}$} & 74.66$_{\pm0.70}$ & 64.79$_{\pm0.99}$ & 12.45$_{\pm2.97}$ & 91.88$_{\pm0.32}$ & 84.91$_{\pm0.38}$ & 6.12$_{\pm0.50}$ \\
    ENGINE~\cite{zhou2025external} & --- & 83.65$_{\pm0.27}$ & 74.75$_{\pm0.52}$ & 11.06$_{\pm0.29}$ & 84.39$_{\pm0.44}$ & 78.12$_{\pm0.10}$ & 7.32$_{\pm1.35}$ & 76.84$_{\pm0.40}$ & 65.85$_{\pm0.21}$ & 11.03$_{\pm0.78}$ & \underline{92.16$_{\pm0.46}$} & 85.65$_{\pm0.10}$ & 6.18$_{\pm0.22}$ \\
    BOFA~\cite{li2026bofa} & --- & 83.82$_{\pm0.32}$ & \underline{75.68$_{\pm0.07}$} & 8.41$_{\pm0.68}$ & \underline{84.47$_{\pm0.43}$} & \underline{78.97$_{\pm0.16}$} & 6.99$_{\pm1.11}$ & \underline{76.98$_{\pm0.91}$} & \underline{66.11$_{\pm0.22}$} & \underline{10.21$_{\pm0.93}$} & 92.14$_{\pm0.49}$ & \underline{86.46$_{\pm0.10}$} & \underline{5.10$_{\pm0.44}$} \\
    \midrule
    Upper-bound & --- & - & 80.91$_{\pm0.06}$ & - & - & 81.46$_{\pm0.12}$ & - & - & 73.46$_{\pm0.38}$ & - & - & 94.97$_{\pm0.21}$ & - \\
    DPT4CL (ours) & --- & \textbf{84.93$_{\pm0.20}$} & \textbf{76.08$_{\pm0.70}$} & \underline{6.70$_{\pm0.80}$} & \textbf{85.08$_{\pm0.57}$} & \textbf{79.14$_{\pm0.25}$} & \textbf{6.65$_{\pm0.67}$} & \textbf{78.86$_{\pm0.06}$} & \textbf{66.61$_{\pm0.14}$} & \textbf{9.40$_{\pm1.89}$} & \textbf{92.82$_{\pm0.61}$} & \textbf{86.83$_{\pm0.51}$} & \textbf{3.46$_{\pm0.30}$} \\
    \bottomrule
\end{tabular}
}
\end{table*}

\subsection{Benchmark Comparison}
In this section, we evaluate DPT4CL against various representative and state-of-the-art methods on four widely used benchmarks under a 10-task class-incremental setting. The mean and standard deviation of Avg., Last, and FM over three random seeds are reported in Table~\ref{tab:main_results}. Overall, DPT4CL achieves the best performance on 11 out of 12 evaluation metrics and the second-best performance on the remaining one, demonstrating its strong generalization and anti-forgetting capability across diverse CL scenarios.

\subsubsection{Comparison with traditional CL methods}
We first compare DPT4CL with traditional CL methods, including LwF, iCaRL, DER, and TagFex. Although all compared methods adopt the same pre-trained CLIP ViT-B/16 backbone for a fair comparison, DPT4CL substantially outperforms these traditional methods across all datasets and metrics. These results indicate that directly transferring conventional CL strategies to a CLIP-based backbone is still insufficient for class-incremental learning. In contrast, DPT4CL effectively exploits the semantic guidance of CLIP through textual prompt tuning, highlighting the importance of language-driven semantic adaptation in CL scenarios.

\subsubsection{Comparison with prompt-based methods}
We then compare DPT4CL with representative prompt-based CL methods. As shown in Table~\ref{tab:main_results}, CoOp even underperforms Continual-CLIP on several datasets, especially in terms of Last accuracy. For instance, compared with Continual-CLIP, CoOp yields Last accuracy changes of $-3.48\%$, $-5.72\%$, and $-2.29\%$ on CIFAR-100, CUB-200, and UCF-101, respectively, indicating that simply optimizing shared prompts can introduce severe classifier bias and degrade the model's performance on previously learned tasks. In contrast, DPT4CL consistently improves over Continual-CLIP across all datasets, with average changes of $+10.89\%$ in Avg. accuracy, $+11.93\%$ in Last accuracy, and $-3.51\%$ in FM over the four datasets. DPT4CL also consistently surpasses prompt pool methods, including AttriCLIP, L2P, DualPrompt, and CODA-Prompt. Unlike these methods, which rely on query--key matching to retrieve task-specific prompts during inference, DPT4CL decouples the learnable textual prompt into task-shared and class-specific components. This design avoids explicit prompt retrieval while preserving cross-task transferable knowledge and category-level discrimination, demonstrating the effectiveness of DPT4CL as a new prompt-based CL paradigm.

\subsubsection{Comparison with CLIP-based SOTA methods}
We further compare DPT4CL with recent CLIP-based SOTA methods, including adapter-based methods such as CLAP4CLIP and ENGINE, and the LoRA-based method BOFA. As shown in Table~\ref{tab:main_results}, DPT4CL achieves superior performance over these methods, particularly in terms of FM. For example, DPT4CL obtains the lowest FM of 3.46\% on UCF-101, demonstrating its strong ability to mitigate catastrophic forgetting. This further validates the effectiveness of the IB-based decoupling strategy in preserving task-shared consensus knowledge while learning class-specific semantics. Since DPT4CL performs adaptation only in the input-space, it is potentially complementary to feature-space adaptation strategies such as adapter-based and LoRA-based fine-tuning. For example, CLAP4CLIP could further benefit from replacing its textual prompt tuning component with more advanced designs. This leaves room for future work on jointly optimizing DPT4CL with feature-space adaptation for CLIP-based continual learning.

\subsubsection{Comparison with rehearsal-based methods}
Finally, compared with rehearsal-based methods, such as LwF, iCaRL, DER, TagFex, and CLAP4CLIP, DPT4CL is rehearsal-free. Despite using no stored exemplars from previous tasks, it achieves markedly better performance on most benchmarks. This highlights the efficiency of DPT4CL in resource-constrained scenarios, where storing or revisiting previous data may be infeasible due to memory, privacy, or deployment constraints.

\begin{table}[t]
\centering
\caption{Component ablation study on CIFAR-100.}
\label{tab}
\resizebox{\columnwidth}{!}{%
\begin{tabular}{lccc}
\toprule
Variants & Avg.~($\uparrow$) & Last~($\uparrow$) & FM~($\downarrow$) \\
\midrule
Baseline (CoOp) & 77.22$_{\pm0.39}$ & 64.89$_{\pm2.23}$ & 18.98$_{\pm2.30}$ \\
Only task-shared prompt & 81.10$_{\pm0.69}$ & 70.78$_{\pm0.36}$ & 9.29$_{\pm0.89}$ \\
Only class-specific prompts & 82.77$_{\pm0.39}$ & 73.73$_{\pm0.28}$ & 7.64$_{\pm0.68}$ \\
DPT4CL w/o $\mathcal{L}_{KD}$ & 83.30$_{\pm0.63}$ & 73.26$_{\pm0.79}$ & 8.41$_{\pm0.57}$ \\
DPT4CL w/o $\mathcal{L}_{ort}$ & 84.55$_{\pm0.24}$ & 75.56$_{\pm0.26}$ & 7.46$_{\pm1.03}$ \\
DPT4CL & \textbf{84.93$_{\pm0.20}$} & \textbf{76.08$_{\pm0.70}$} & \textbf{6.70$_{\pm0.80}$} \\
\bottomrule
\end{tabular}%
}
\vspace{-3mm}
\end{table}

\subsection{Further Analysis}
\subsubsection{Ablation Study}
To evaluate the contribution of each component in Eq.~(\ref{DPT4CL}), we conduct an ablation study on CIFAR-100. As shown in Table~\ref{tab}, when only applying $\mathcal{L}_{CE}$, the framework degenerates to the baseline CoOp, yielding suboptimal average accuracy and severe forgetting. The variant \textit{Only task-shared prompt} learns only the task-shared prompt distribution through the IB objective $\mathcal{L}_{IB}$ on top of $\mathcal{L}_{CE}$, while \textit{Only class-specific prompts} introduces only class-specific prompts and the corresponding orthogonalization loss based on $\mathcal{L}_{CE}$. Moreover, \textit{DPT4CL w/o $\mathcal{L}_{KD}$} and \textit{DPT4CL w/o $\mathcal{L}_{ort}$} are used to investigate the effects of removing $\mathcal{L}_{KD}$ and $\mathcal{L}_{ort}$ from the complete framework, respectively. With all components jointly optimized, DPT4CL achieves the best Avg. accuracy of 84.93\%, Last accuracy of 76.08\%, and FM of 6.70\%, validating the complementary contributions of these components.

\begin{table}[t]
\centering
\caption{Performance varies with length $L$ of prompt on CIFAR-100 and UCF-101.}
\label{tab:prompt_length}
\setlength{\tabcolsep}{2.5pt}
\renewcommand{\arraystretch}{1.05}
\resizebox{\columnwidth}{!}{%
\begin{tabular}{@{}c ccc ccc@{}}
\toprule
\multirow{2}{*}{Length of prompt}
& \multicolumn{3}{c}{CIFAR-100}
& \multicolumn{3}{c}{UCF-101} \\  
\cmidrule(lr){2-4} \cmidrule(lr){5-7}
& Avg. $\uparrow$ & Last $\uparrow$ & FM $\downarrow$
& Avg. $\uparrow$ & Last $\uparrow$ & FM $\downarrow$ \\ 
\midrule
1 & 84.39$_{\pm0.41}$ & 75.50$_{\pm0.23}$ & 6.79$_{\pm0.60}$
& 92.58$_{\pm0.52}$ & 86.49$_{\pm0.33}$ & 4.12$_{\pm0.16}$ \\ 
2 & \textbf{84.93$_{\pm0.17}$} & \textbf{76.08$_{\pm0.57}$} & \textbf{6.70$_{\pm0.66}$}
& 92.82$_{\pm0.50}$ & 86.83$_{\pm0.42}$ & \textbf{3.46$_{\pm0.24}$} \\ 
3 & 84.85$_{\pm0.20}$ & 76.06$_{\pm0.25}$ & 7.83$_{\pm0.75}$
& \textbf{93.07$_{\pm0.18}$} & \textbf{87.31$_{\pm0.67}$} & 3.55$_{\pm0.21}$ \\ 
4 & 84.86$_{\pm0.21}$ & 75.94$_{\pm0.24}$ & 6.94$_{\pm0.39}$
& 92.79$_{\pm0.35}$ & 86.79$_{\pm0.98}$ & 3.57$_{\pm0.30}$ \\ 
\bottomrule
\end{tabular}%
}
\vspace{-3mm}
\end{table}

\begin{figure}[!t]
\centering
\includegraphics[width=0.82\columnwidth]{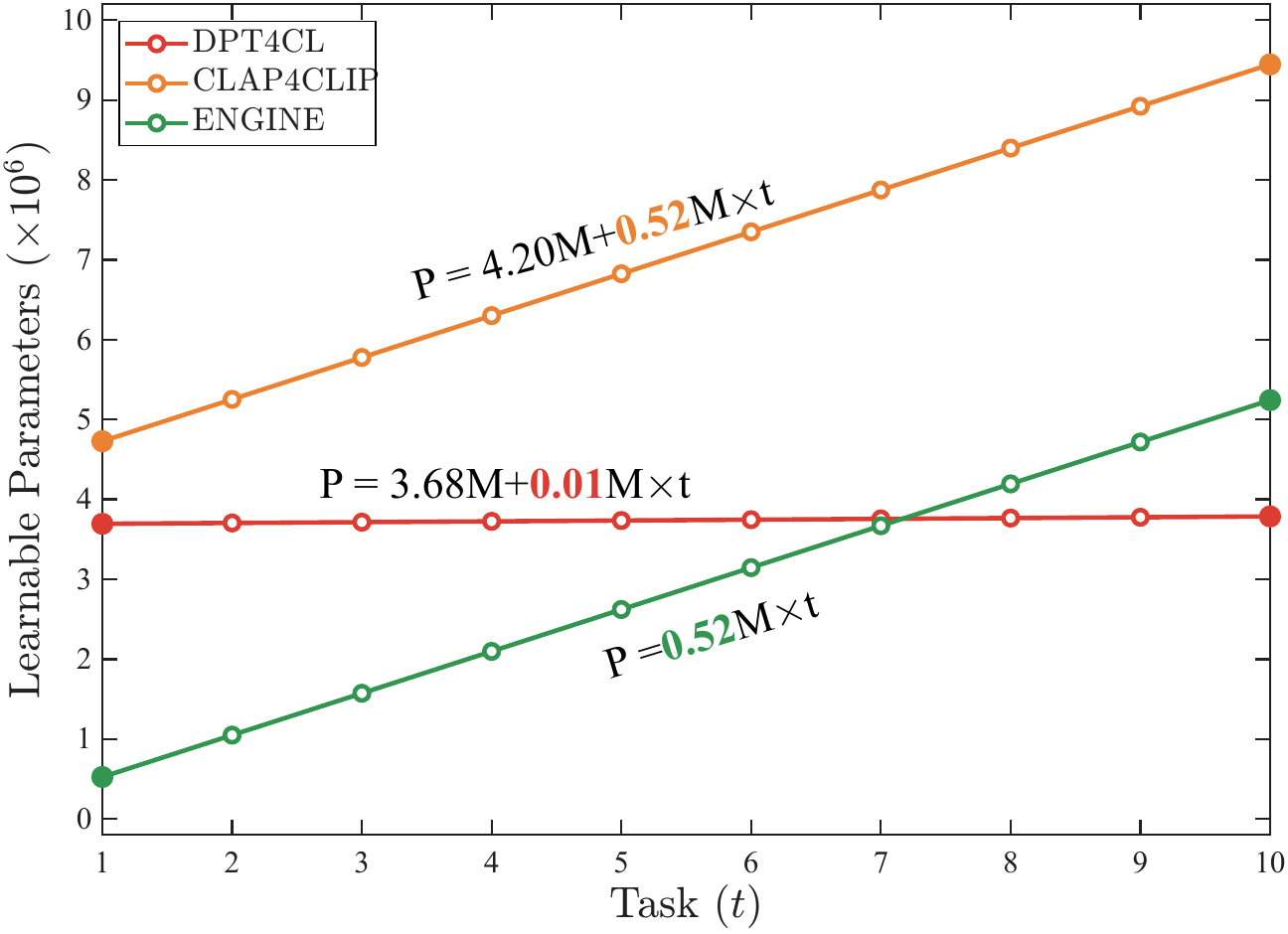}
\caption{Learnable parameter growth with incremental tasks.}
\label{fig:parameter_count}
\vspace{-3mm}
\end{figure}

\subsubsection{Hyperparameter analysis}
We conduct a sensitivity analysis on the prompt length $L$, which denotes the length of both the task-shared prompt and the class-specific prompts. As shown in Table~\ref{tab:prompt_length}, DPT4CL achieves stable performance across different values of $L$ on CIFAR-100 and UCF-101. In particular, $L=2$ yields the best overall trade-off, achieving the best results on CIFAR-100 and the lowest FM on UCF-101, while larger prompt lengths bring no consistent performance gains. Therefore, we set $L=2$ as the default configuration for all benchmarks. Detailed analyses for the remaining loss coefficient hyperparameters ($\alpha$, $\beta$, $\lambda$, and $\eta$) are provided in the Supplementary Material.

\subsubsection{Parameter count analysis}
Fig.~\ref{fig:parameter_count} compares the learnable parameter growth of DPT4CL with CLIP-based SOTA methods on CIFAR-100. With a ViT-B/16 backbone and prompt dimension $d=512$, DPT4CL introduces about 3.68M task-shared parameters from the prompt encoder and classifier. Its incremental parameters only come from class-specific prompts for newly arrived classes. Since each task contains 10 new classes, DPT4CL adds only $L \times d \times 10 = 10{,}240$ parameters per task, i.e., about 0.01M when $L=2$. In contrast, CLAP4CLIP and ENGINE introduce adapter-style modules with $2 \times d^2 = 524{,}288$ parameters per task. Thus, DPT4CL requires only about $1/50$ of their incremental parameters, demonstrating better parameter efficiency and scalability while achieving superior performance in Table~\ref{tab:main_results}.

\begin{figure}[!t]
\centering
\setlength{\tabcolsep}{1pt}
\renewcommand{\arraystretch}{1.0}

\begin{tabular}{@{}c@{\hspace{1mm}}c@{}}
\begin{minipage}[t]{0.47\columnwidth}
\centering
\includegraphics[width=\linewidth]{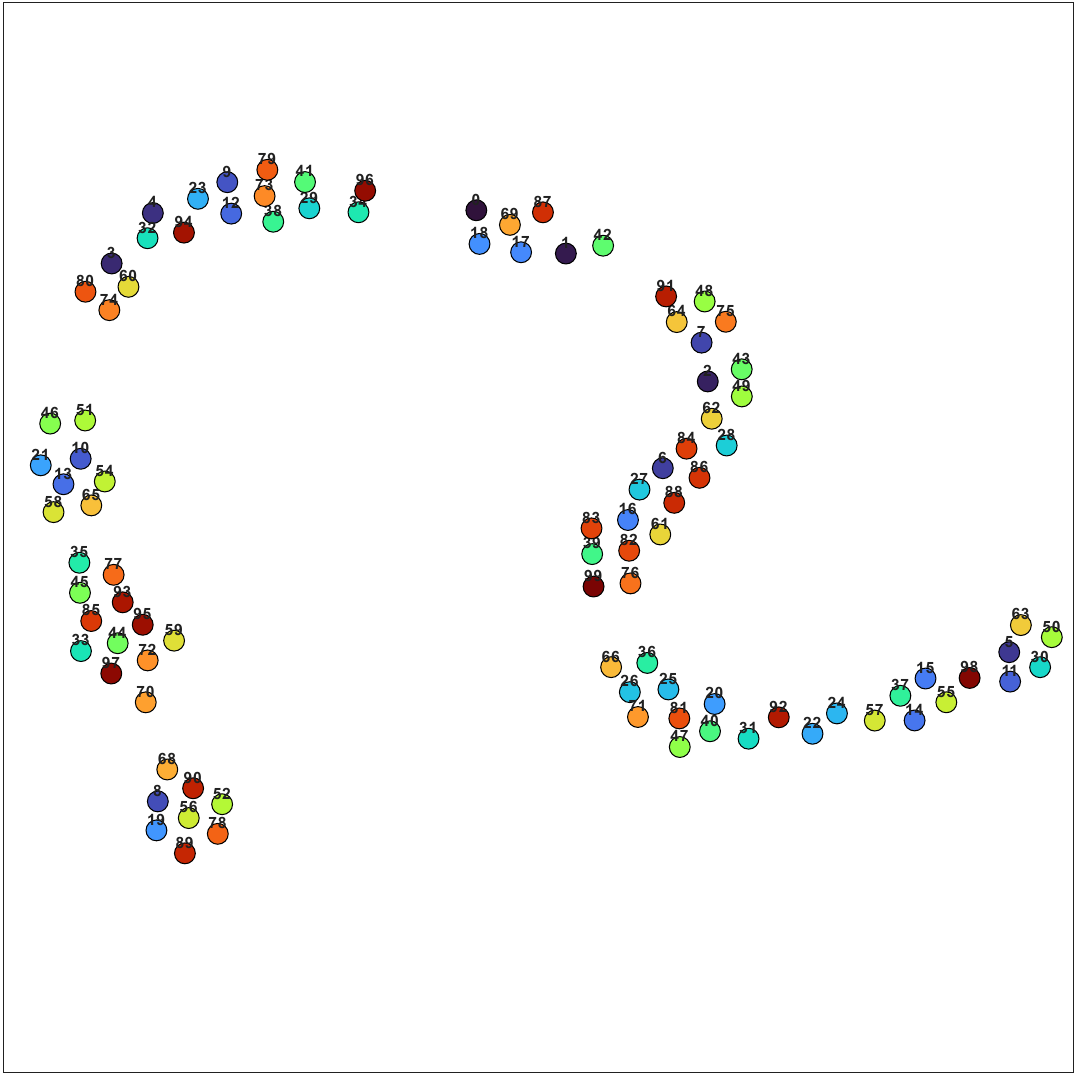}
\\[-1.5mm] 
{\scriptsize (a) DPT4CL task-shared prompt}
\end{minipage}
&
\begin{minipage}[t]{0.47\columnwidth}
\centering
\includegraphics[width=\linewidth]{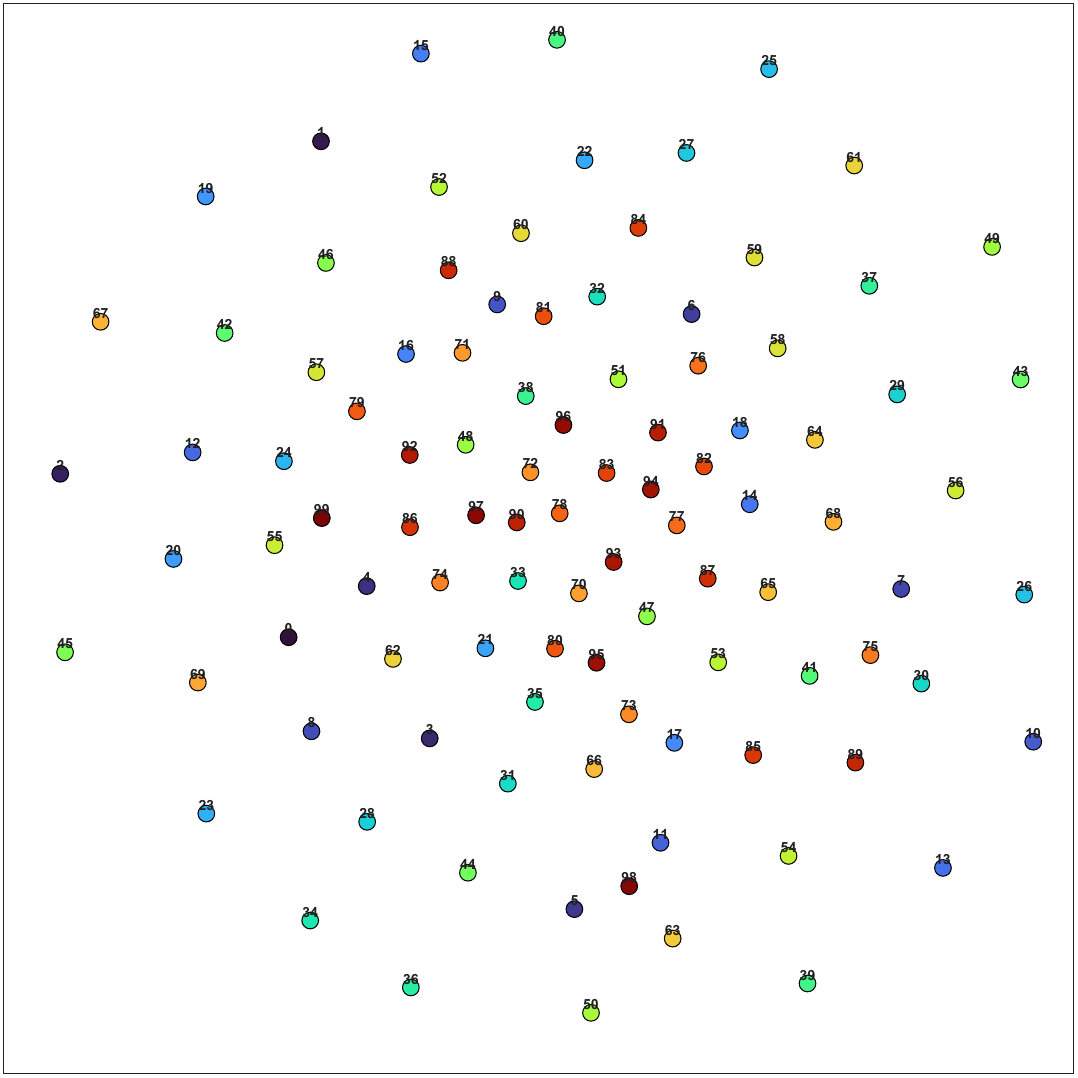}
\\[-1.5mm] 
{\scriptsize (b) DPT4CL concatenated prompts}
\end{minipage}
\tabularnewline[0mm] 
\end{tabular}

\caption{t-SNE visualizations of DPT4CL prompts on CIFAR-100. Different colors denote different categories.}
\label{fig:all_vis}
\vspace{-4mm}
\end{figure}

\begin{figure}[t]
\centering
\includegraphics[width=\columnwidth]{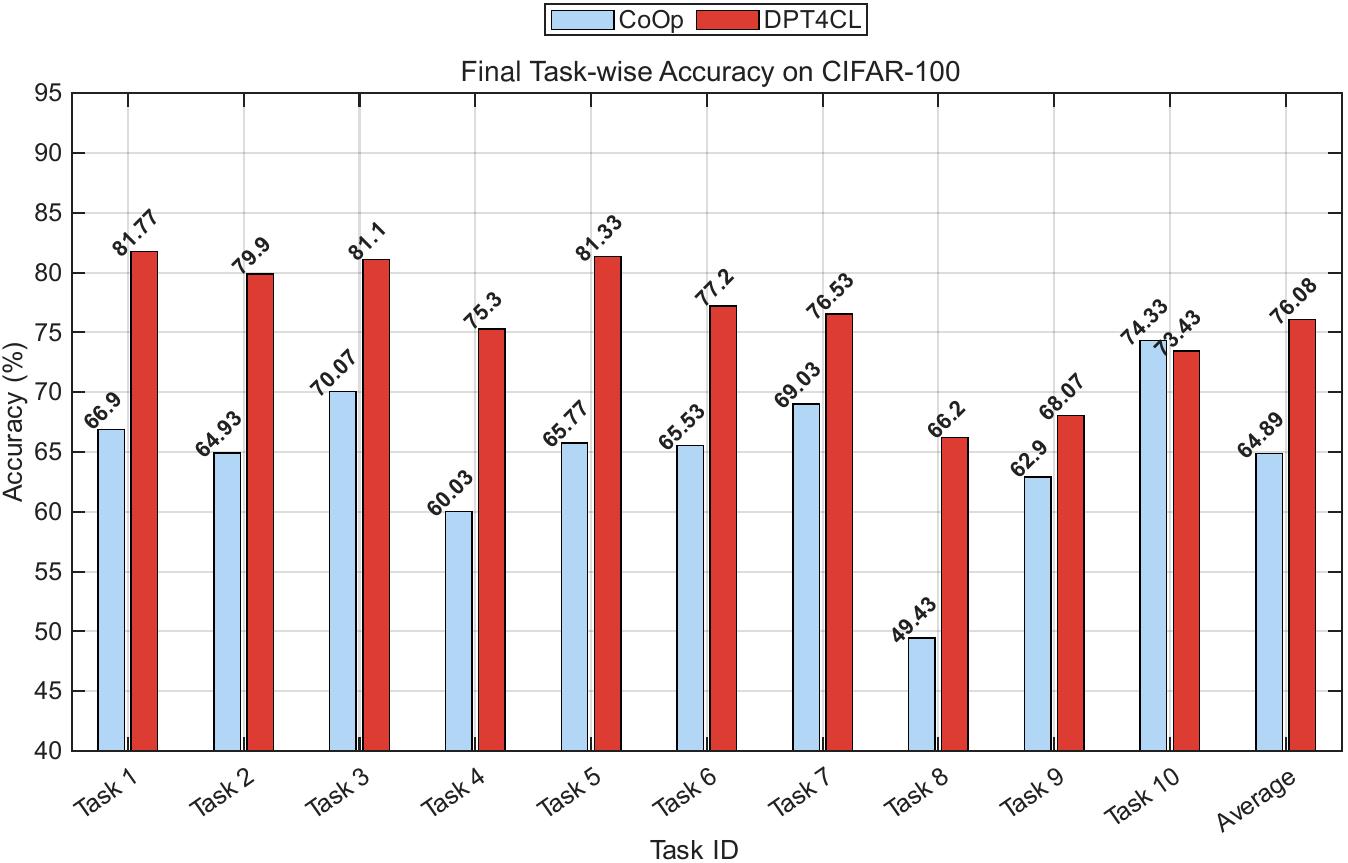}
\caption{Final task-wise accuracy of CoOp and DPT4CL on CIFAR-100 after training all 10 tasks. Average corresponds to the Last metric.}
\label{fig}
\vspace{-4mm}
\end{figure}

\subsubsection{Visualization analysis}
We visualize the learned prompts of DPT4CL on CIFAR-100 using t-SNE. As shown in Fig.~\ref{fig:all_vis}(a), the task-shared prompts form a continuous manifold in the prompt space, indicating that they capture shared knowledge transferred across different tasks. Fig.~\ref{fig:all_vis}(b) shows that the complete prompts obtained by concatenating the task-shared prompt with class-specific prompts exhibit a more discrete distribution with enlarged inter-class margins in the prompt space, suggesting that the class-specific prompts introduce additional category-discriminative information.

To further analyze the classifier bias caused by CoOp-like prompt optimization, we compare the final task-wise accuracy of CoOp and DPT4CL on CIFAR-100 after training all 10 tasks. As shown in Fig.~\ref{fig}, CoOp exhibits clear performance degradation on old tasks, while DPT4CL achieves consistently higher accuracy on most previously learned tasks. These results demonstrate that the proposed decoupled prompt design effectively alleviates classifier bias and better preserves knowledge from previous tasks.

\section{Conclusion}
In this paper, we propose DPT4CL, a rehearsal-free prompt tuning framework for CLIP-based continual learning. Different from existing prompt-based methods, DPT4CL provides a new prompt tuning paradigm for CL by decoupling the prompt design into task-shared and class-specific components, thereby avoiding explicit prompt retrieval while maintaining effective knowledge transfer and discriminative prompt representations across sequential tasks. Extensive experiments on four standard CL benchmarks demonstrate the effectiveness of DPT4CL, which achieves the best performance on 11 out of 12 evaluation metrics and the second-best performance on the remaining one. Moreover, the proposed framework is supported by a unified excess risk analysis from an information-theoretic perspective, providing theoretical insight into its ability to control generalization and forgetting. In addition, its rehearsal-free nature and small incremental parameter cost make it suitable for resource-constrained scenarios. In future work, we will explore the joint optimization of input-space prompt tuning and feature-space adaptation strategies for CL.


\bibliographystyle{IEEEtran}
\bibliography{IEEEabrv, CL-REF}

\end{document}